\documentclass{article}

\usepackage{arxiv}

\usepackage[utf8]{inputenc} 
\usepackage[T1]{fontenc}    
\usepackage{hyperref}       
\usepackage{url}            
\usepackage{booktabs}       
\usepackage{amsfonts}       
\usepackage{microtype}      
\usepackage{amsmath}
\usepackage{graphicx}       
\usepackage{enumitem}       
\usepackage{multicol}       
\usepackage{todonotes}      

\usepackage[ruled,vlined]{algorithm2e}

\newcommand{\orcid}[2]{\href{https://orcid.org/#1}{\includegraphics[scale=0.06]{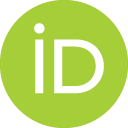}\hspace{1mm} #2}}

\title{Hierarchical Qualitative Clustering: clustering\\
mixed datasets with critical qualitative information\\\thanks{This work was financed by National Funds through the Portuguese funding agency, FCT - Fundação para a Ciência e a Tecnologia, within project UIDB/50014/2020.}}

\author{
    \orcid{0000-0003-2652-3718}{Diogo Seca}\\
    LIAAD - INESC TEC\\
    Porto, Portugal\\
    \texttt{jose.d.seca@inesctec.pt}\\
\And
    \orcid{0000-0002-9081-2728}{João Mendes-Moreira}\\
    LIAAD - INESC TEC, \\
    DEI - FEUP, Universidade do Porto\\
    Porto, Portugal\\
    \texttt{jmoreira@fe.up.pt}\\
\AND
    \orcid{0000-0002-4802-7558}{Tiago Mendes-Neves}\\
    LIAAD - INESC TEC\\
    Porto, Portugal\\
    \texttt{tiago.m.neves@inesctec.pt} \\
\And
    \orcid{0000-0002-8414-5826}{Ricardo Sousa}\\
    LIAAD - INESC TEC\\
    Porto, Portugal\\
    \texttt{ricardo.t.sousa@inesctec.pt}\\
}

\begin{document}
\maketitle
\begin{abstract}

Clustering can be used to extract insights from data or to verify some of the assumptions held by the domain experts, namely data segmentation. 
In the literature, few methods can be applied in clustering qualitative values using the context associated with other variables present in the data, without losing interpretability. Moreover, the metrics for calculating dissimilarity between qualitative values often scale poorly for high dimensional mixed datasets.
In this study, we propose a novel method for clustering qualitative values, based on Hierarchical Clustering (HQC), and using Maximum Mean Discrepancy. HQC maintains the original interpretability of the qualitative information present in the dataset. We apply HQC to two datasets. Using a mixed dataset provided by Spotify, we showcase how our method can be used for clustering music artists based on the quantitative features of thousands of songs. In addition, using financial features of companies, we cluster company industries, and discuss the implications in investment portfolios diversification.

\end{abstract}
\vspace{3mm}

\begin{multicols}{2}
\section{Introduction}

Among data mining methodologies, it is common to find the Business Understanding and Data Understanding as the phases with the most priority. In CRISP-DM, Business Understanding is the first phase of the model process, and Data Understanding is the second \cite{shearer2000}. In SCRUM-DM, Business Understanding is the first task of the project, and Data Understanding is performed every one to four weeks \cite{nogueira2014}. Furthermore, practitioners often revisit these phases due to the importance they have in the context of a data science project.

Indeed, existing domain knowledge can be a deciding factor for the success of a data mining project. In the cases where the domain experts are not available or accessible, practitioners can use unsupervised learning techniques to extract insights about the nature of the data. Alternatively, the same unsupervised learning techniques are useful in verifying some of the assumptions made by the domain experts, namely data segmentation through the use of Clustering techniques.

The focus of this study is the clustering of qualitative variables using as much information as possible from the data while still maintaining interpretability.

\subsection{Contributions}

The contributions of this paper are threefold:
\begin{enumerate}[leftmargin=6mm]
\item brief review of dissimilarity and statistical distance measures for the clustering of qualitative values;
\item introduction of a new algorithm for the hierarchical clustering of qualitative values;
\item demonstration and visualization of the suggested procedure on two real datasets.
\end{enumerate}

\subsection{Paper Structure}

The remainder of the paper's organization is as follows. Section 2 briefly reviews related work and the background required to understand the following sections of the paper. In Section 3, we present a novel procedure called "Hierarchical Qualitative Clustering" and discuss its most relevant details. In Section 4, we present two use case scenarios using real datasets, resulting in two interpretable visualizations. Section 5 concludes the results with practical information for machine learning practitioners. Section 6 points to open research questions emerging from this work.
\section{Background}

\subsection{Clustering}

Clustering aims at partitioning data into groups to organize data into a more meaningful form. Two of the most prominent types of clustering are Partitional Clustering and Hierarchical Clustering. 

The goal of Partitional Clustering is to maximize the homogeneity within the clusters, and heterogeneity between the clusters. Clusters do not overlap and each instance can only belong to one cluster. Perhaps the most famous among Paritional Clustering algorithms is k-means, due to its efficiency. k-means is only able to deal with quantitative data. In order to use quantitative distance metrics in a mixed dataset, we first need to either convert the qualitative variables to quantitative or drop them from the dataset.

Unlike other clustering algorithms such as k-means, Hierarchical Clustering does not require the specification of the number of clusters. Hierarchical Clustering focuses on building a hierarchical structure of clusters. In these methods, a cluster can contain clusters lower in the hierarchy and can belong to clusters higher in the hierarchy. Hierarchical Clustering requires the specification of a measure of dissimilarity between clusters. Moreover, Hierarchical Clustering can either be Divisive, following a top-down approach, or Agglomerative, following a bottom-up approach. In Hierarchical Divisive Clustering, the instances start as one single cluster, which is recursively split as we move down the hierarchy. Contrastingly, in Hierarchical Agglomerative Clustering, each instance starts in its singleton cluster. The clusters are then iteratively paired until one global cluster remains \cite{hastie2009}.

\subsection{Data-types}

Variables can either be quantitative or qualitative. In real datasets and applications, we can frequently describe the observations by a combination of both quantitative and qualitative variables. Such datasets are known as mixed or mixed-type data. However, most distance metrics and distance-based clustering methods work either with quantitative-only or qualitative-only data \cite{velden2019}. 

Among qualitative variables, ordinal variables contain information about the order of magnitude of the qualitative value, and can, therefore, be converted into quantitative, e.g., 'Low' becoming 1, 'Medium' becoming 3, and 'High' becoming 9. In the particular case in which the ordinal variable represents bins, it can also be converted into the mean, median, or midpoint of the bin. 

Another type of qualitative variable are nominal variables. Nominal variables cannot be ordered in magnitude like ordinal variables \cite{rezankova2009}. Some of the most popularized algorithms for transforming nominal variables into quantitative features are:
\begin{itemize}[leftmargin=6mm]
\item one-hot encoding, also known as dummy coding, in which a new binary variable is added for each unique qualitative value;
\item hashing, similar to one-hot-encoding but results in fewer dimensions, and there is some loss of information due to collisions;
\item embeddings, a technique frequently used in Natural Language Processing for mapping words and sentences into vectorial spaces;
\item geocoding, i.e., transforming names of physical locations into the coordinates of latitude and longitude.
\end{itemize}

An adaptation of k-means, k-modes is used for clustering qualitative variables exclusively. The method defines its clusters based on the number of matching qualitative values between instances. One way quantitative variables can be discretized is by using the supervised entropy-based Minimum Description Length Principle (MDLP) algorithm proposed by Fayyad and Irani~\cite{fayyad1993}.

\subsection{Clustering mixed data}

Velden et al. distinguish three ways of clustering mixed-type data: (1) by converting data to the same scale before clustering using either qualitative-based distances or quantitative-based distances; (2) by using specific distance measures for mixed data; (3) by using specific clustering methods designed for mixed data \cite{velden2019}.

An example of the latter is the k-prototypes algorithm, which combines both k-modes and k-means and is, therefore, able to cluster mixed-type data \cite{huang1997}.

In an attempt to speed-up the modeling process, some machine learning practitioners often choose a more straightforward solution: to drop the qualitative variables and use quantitative-only methods. Such a practice is not unreasonable, since opting for commonly proposed solutions such as one-hot encoding of the qualitative variables could significantly increase dimensionality. One-hot encoding is especially problematic for nominal variables with a large number of unique values. Such a practice could lead to worse modeling. For more info, see the curse of dimensionality \cite{domingos2012}. On the other hand, dropping the qualitative variable could cause the loss of relevant information. Moreover, the process of converting qualitative values into quantitative data also causes loss of interpretability in the data, and hence, loss in interpretability of models learned from the data.

\subsection{Dissimilarity Measures}

In this subsection, we focus on exploring the literature on dissimilarity measures, with usability in the clustering of qualitative values. Dissimilarities for mixed-type data are beyond the scope of this study, and so we will not be exploring commonly used metrics such as the Gower dissimilarity. For more info, see \cite{gower1971}.

For qualitative data, one of the most common distance measures is the Szymkiewicz-Simpson metric, commonly known as the overlap metric:

$$d_{\text{overlap}}(x,y)={\frac {|x\cap y|}{\min(|x|,|y|)}} \;  ,$$

in which $x$ and $y$ represent vectors of two instances, composed only by qualitative values.
 
Other commonly used distance, based on the Jaccard coefficient, is the Jaccard distance:

$$ d_{\text{Jaccard}}(x,y) = 1 - \frac{|x \cap y|}{|x \cup y|} \;  .$$
 
Several partitional and hierarchical clustering algorithms have adopted the Jaccard distance, namely k-modes \cite{huang1998}, and LIMBO \cite{andritsos2004}.

\subsubsection{Context-sensitive measures}

While some studies focus only on the available qualitative variables \cite{kimes2017} for clustering, in this study, we hypothesize that there is potentially useful information in the distributions of quantitative variables frequently associated with each qualitative value.

Indeed, similarity measures for qualitative variables can be grouped into context-free and context-sensitive. Most commonly used measures of distance are context-free. Context-free distances between qualitative variables ignore the dependency effects from other variables. Conversely, context-sensitive measures make use of information contained in other variables \cite{alamuri2014}.

Le and Ho \cite{le2005} proposed a dissimilarity measure for qualitative data based estimating the Kullback-Leibler divergence between the conditional distributions of the instances values.




Other possible similarity measures based on correlation are: Distance Correlation \cite{szekely2014}, Canonical Correspondence Analysis \cite{hardoon2004} and the RV‐coefficient \cite{robert1976}. These statistics for two-sample tests are implemented in the Python package \emph{hyppo}, along with the estimation of p-values using Bootstrap. Note that correlation-based similarity measures are not suitable in scenarios where the attributes are highly independent.

Ienco et al. introduced DIstance Learning in Categorical Attributes (DILCA), a method to compute a context-based distance between values of a qualitative variable \cite{ienco2012}. The authors apply this technique to the hierarchical clustering of qualitative data. The dissimilarity measure used takes into account other qualitative variables in the context. However, this method is not able to deal with information in quantitative variables - it requires the discretization of continuous variables. Ienco et al. use the discretization method MDL described in \cite{fayyad1993}, the same method used by k-means. Contrastingly, in our study, we do not require the discretization of variables. We can use quantitative variables, including discrete and continuous, in order to extract the context required to calculate the dissimilarity between qualitative values.

\subsubsection{Statistical distances}

Khorshidpour et al. propose using the Kullback-Leibler divergence in order to compute the dissimilarity between probability distributions \cite{khorshidpour2010}. In other words, the dissimilarity between two qualitative values is computed as the Kullback-Leibler divergence between the two empirical conditional distributions. Evidence suggests that using this method improves the generalization of nearest neighbors, in classification problems \cite{khorshidpour2010}. Given two probability distributions $P$ and $Q$, the Kullbalck-Leibler divergence can be defined as:

$$ d_{\text{KL}}(P, Q)=\int p(x)\log \left({\frac {p(x)}{q(x)}}\right) \, dx \; ,$$

in which $p$ and $q$ denote the probability densities of $P$ and $Q$, respectively.

However, the Kullback-Leibler divergence cannot be considered a distance due to not being symmetric. Given two objects $P$ and $Q$, the symmetry condition holds if: 
$$ d(P, Q) = d(Q, P) .$$

Jorge et al. proposed CAREN-DR, an algorithm for subgroup discovery using Distribution Rules (DR) \cite{jorge2006}. Subgroup discovery is a problem akin to clustering. However, while in clustering, a single cluster cannot be evaluated without considering all the other clusters, in subgroup discovery, we can identify groups that are interesting regardless of other potential groups around \cite{lucas2007}. CAREN-DR aims to discover distribution rules that define subgroups within the dataset, whose target distribution is statistically different from the default target distribution. Instead of using the Kullback-Leibler divergence, CAREN-DR uses the Kolmogorov-Smirnov test for understanding whether the same distribution could have originated the two samples observed. The default distribution, i.e., the distribution of reference, is the one obtained with all the values of y for the whole dataset. CAREN-DR can be used in descriptive data mining tasks with the advantage of avoiding the pre-discretization of the quantitative variable of interest. The Kolmogorov-Smirnov statistic can be defined as:

$$d_{\text{KS}}(P, Q)=\sqrt{\frac{nm}{n+m}}\sup _{x}|F_P(x)-F_Q(x)| \;  ,$$

in which $P$ and $Q$ denote two probability distributions, and $F_P$ and $F_Q$ denote its cumulative probability distributions respectively.
 
Currently, there are more powerful alternatives to the Kolmogorov-Smirnov, among which stands the Anderson-Darling (AD) test. When the two statistical then tests are compared, studies show that the AD test is more sensitive in differences in shift, scale, and symmetry w.r.t distributions. Moreover, the AD test is more sensitive to differences in densities near the tails of the distributions. Engmann and Cousineau also find that the AD test "requires less data than the KS test to reach sufficient statistical power" \cite{engmann2011}. Based on the AD statistic defined by Pettitt \cite{pettitt1976}, we compute the AD distance as:

$$ d_{\text{AD}}(P, Q)=\frac{nm}{n+m} \int \frac{(P(x) - Q(x))^2}{H_N(x)(1-H_N(x))} \, dH_N(x)\;  ,$$

where $P$ and $Q$ denote the two probability distributions of samples $X$ and $Y$. $n$ and $m$ denote the sample size of $X$ and $Y$ respectively. $H_N$ is the combined probability distribution that arises from combining samples $X$ and $Y$. $H_N$ can be computed as:

$$H_N(x) = \frac{nF(x) + mP(x)}{n+m}. $$ 
 
The KS and AD tests defined above are useful when working with univariate distributions. The most typical definitions of the tests contemplate the test to be performed along one dimension. For clustering or subgroup discovery, this means that the user has to pick one dimension on which to perform the test. More commonly, the choice of the dimension is the target variable. The distributions compared are subsets of the dataset, defined by a distribution rule. However, in extracting the context, it would be useful to use, not just one variable, but all of the available quantitative variables. For that, we would require an N-dimensional two-sample test. However, in contrast to the univariate case, the probability distribution functions of statistics Kolmogorov-Smirnov and Anderson-Darling in the multivariate case are not distribution-free \cite{baksajev2015}. Bakšajev and Rudzkis note that this problem can be overcome by using the Rosenblatt transformation, as shown in \cite{rosenblatt1952}. However, the same authors also note considerable difficulties in computing the statistical distances mentioned using this process \cite{baksajev2015}.

\subsubsection{Maximum Mean Discrepancy}

One way we can quantify the difference between two multivariate distributions is to use Maximum Mean Discrepancy (MMD) \cite{gretton2012}. MMD is a faster alternative to tests like Kolmogorov-Smirnov, Cramer von Mises, and Anderson-Darling. MMD has become popular due to its use in problems with a high number of dimensions. One particular study bases itself on the theory behind MMD and energy statistics to build a new clustering method, called kernel k-groups. The method tends to outperform K-means, especially in higher dimensions \cite{franca2017}. MMD is closely related to energy statistics and in particular energy distance.

The simplified intuition behind MMD is to compare the means of both distributions. However, this could cause us to miss certain differences in the distributions, such as variance, skewness, kurtosis, and other more exotic characteristics. So instead of comparing the means of the two raw distributions, we learn distinguishing features, and compare the means of those features. Given two samples $X$ and $Y$ from distributions $P$ and $Q$ respectively, the main formula for the MMD distance is as follows:

$$ d_{\text{MMD}}(P, Q) = \sup_{f \in F}|\ E_{X \sim P}[f(X)] - E_{Y \sim Q}[f(Y)]\ |_H  \;  ,$$

in which $H$ is the reproducing kernel Hilbert space which maps the features to the original variables. The unbiased MMD between two samples can be simplified to:

\begin{equation*}
    \centering
    \begin{split}       
    d_{\text{MMD}}(P, Q)^2 &= \frac{2}{n(n-1)} \sum_{i=1}^n \sum_{j=i+1}^n k(x_i,x_j) \\
                           &+ \frac{2}{m(m-1)} \sum_{i=1}^m \sum_{j=i+1}^m k(y_i,y_j) \\
                           &- \frac{2}{mn} \sum_{i=1}^n \sum_{j=1}^m k(x_i,y_j) \; ,   
    \end{split} 
\end{equation*}

in which $k(x, x')$ denotes the kernel function.

As with other energy statistics, one way to estimate the p-value for the hypothesis tests is by using the Bootstrap on the aggregated data, following Arcones and Gine (1992) \cite{arcones1992}. 
The null hypothesis, in this case, is that the two samples came from the same distribution.

In the remainder of this study, we will be using MMD as the default distance metric.
s\section{Hierarchical Qualitative Clustering}

In this section, we will explain the procedure required for the hierarchical clustering of qualitative variables in mixed-type data.

The method we present is similar to Hierarchical Agglomerative Clustering (HAC). At each iteration, the clusters that are joined are those that are most similar, i.e., those clusters whose distance is smallest. In contrast to HAC, we do not start with singleton clusters. Instead, the instances are initially grouped by qualitative value. Each group forms a cluster that can be identified by a qualitative value. Intuitively, qualitative variables can be seen as interpretable clues for the formation of the initial clusters. Thereon, given the several quantitative variables in the dataset, each cluster's distribution is compared with that of other clusters.

\subsection{Definition of distance}

Given that a dataset has several quantitative variables, the statistical distance must be able to perform N-dimensional two-sample tests. The distance measure used for this method between clusters is Maximum Mean Discrepancy (MMD). Given a selected qualitative variable $C$ for clustering, instead of using calculating the distances directly from the qualitative values $c_a$ and $c_b$ themselves, we calculate the MMD distance $d$ from the two distributions of quantitative variables $X$ conditional on the qualitative values $c_a$ and $c_b$ respectively. The distance can be written as:

$$d(c_a, c_b) = d_{MMD}(\ X|C=c_a,\ X|C=c_b\ )\; .$$

Theoretically, the distance should always be positive. However, the squared MMD can sometimes be computed as less than zero, due to Q overfitting P~\cite{kim2016}. This issue can be avoided by providing a sufficiently large enough sample.  However, when negative values occur for the squared MMD, the distance is considered to be 0.

\subsection{Kernel function and Standardization}

The MMD distance requires the computation of a kernel. Based on the work laid by Gretton et al.~\cite{gretton2012}, the kernel function of choice was the Gaussian Radial Basis Function (RBF), which is defined as: 

$$k_{\text{RBF}}(\mathbf {x} ,\mathbf {x'} )=\exp(-\gamma \|\mathbf {x} -\mathbf {x'} \|^{2})\; ,$$

in which $\gamma$ typically takes the default value of $ {\tfrac {1}{2\sigma ^{2}}}$ and $\sigma$ denotes the sample standard deviation. 

Normalizing the data before feeding it to a learning algorithm is a good practice in most cases. This practice is especially important before the calculation of the kernel matrix using RBF. By standardizing the quantitative variables, we make sure that each quantitative dimension has a variance of around 1.0 and contributes similarly to the L2 norm contained in the RBF kernel.

\subsection{Method description}

Algorithm~\ref{alg:hqc} describes an abridged version of the Hierarchical Qualitative Clustering (HQC) method. The algorithm proposed, HQC, has two main differences from HAC:
\begin{enumerate}[leftmargin=6mm]
\item Clusters are not singleton clusters, but clusters with more than one instances. In particular, each particular cluster contains instances that have the same categorical value(s).
\item The linkage between two clusters is Maximum Mean Discrepancy between the distributions of each of the clusters.
\end{enumerate}


\begin{algorithm}[H]
\caption{Hierarchical Qualitative Clustering}\label{alg:hqc}
\SetAlgoLined
\KwData{Dataset $D$, containing a qualitative variable $Q$, and qualitative variables $X$}
\KwResult{List of initial clusters $initialc$ and a linkage list $linkage$}
\Begin{
    $initialc \leftarrow$ list of distinct $q \in Q$\;
    $linkage \leftarrow$ empty list of tuples\;
    $n \leftarrow$ size of $initialc$\;
    $dissim \leftarrow$ empty matrix, $n \times n$\;
    \For{$c1 \in initialc$}{
        \For{$c2 \in initialc$}{
            $dissim[c1,c2] \leftarrow d_\text{MMD}(X|c1, X|c2)$\;
        }
    }
    \While{$dissim.size \neq 1$}{
        $c1, c2 \leftarrow argmin(dissim)$\;
        $c_\text{new} \leftarrow c1 \cup c2$\;
        $d_\text{new} \leftarrow d_\text{MMD}(X|c1, X|c2)$\;
        $linkage.append((c1, c2, c_\text{new}, d_\text{new}))$\; 
        $dissim.drop(c1)$\;  
        $dissim.drop(c2)$\;  
        $dissim.add(c_\text{new})$\;
        \For{$c \in dissim, c \neq c_\text{new}$}{
            $d \leftarrow d_\text{MMD}(X|c, X|c\text{new})$\;
            $dissim[c, c_\text{new}] \leftarrow d$\;
            $dissim[c_\text{new}, c] \leftarrow d$\;
        }
    }
}
\end{algorithm}

The $linkage$ will allow the user to view the links between clusters. The last entry describes the link that formed the final cluster. The user can choose to cut the list short and keep clusters below a certain threshold, e.g. clusters below 0.4 dissimilarity.

The algorithm presented is a simplified version. To view and experiment with the actual code used, visit \url{https://github.com/diogoseca/qualitative-clustering}.
\section{Use Cases}

\begin{table*}[t]
    \centering
    \begin{tabular}{llllll}
    \toprule
    ID & Child 1 & Child 2 &  Distance &   Size &                                         Artists \\
    \midrule
    30         &      19 &      12 &  0.147 &    663 &                     \{Stevie Wonder, Fleetwood Mac\} \\
    31         &      14 &       2 &  0.160 &    954 &                   \{Ella Fitzgerald, Frank Sinatra\} \\
    32         &      30 &      23 &  0.169 &    911 &      \{Stevie Wonder, Fleetwood Mac, Grateful Dead\} \\
    33         &      32 &      28 &  0.182 &   1136 &  \{Stevie Wonder, Fleetwood Mac, Grateful Dead, ... \\
    34         &      21 &       4 &  0.190 &    779 &                      \{The Who, The Rolling Stones\} \\
    35         &      34 &      27 &  0.186 &   1005 &           \{The Who, The Rolling Stones, The Kinks\} \\
    36         &      31 &      11 &  0.196 &   1363 &      \{Ella Fitzgerald, Frank Sinatra, Dean Martin\} \\
    37         &      36 &      22 &  0.202 &   1612 &  \{Ella Fitzgerald, Frank Sinatra, Dean Martin, ... \\
    38         &      33 &      20 &  0.210 &   1400 &  \{Stevie Wonder, Fleetwood Mac, Grateful Dead, ... \\
    39         &      38 &       3 &  0.210 &   1925 &  \{Stevie Wonder, Fleetwood Mac, Grateful Dead, ... \\
    40         &      29 &      16 &   0.228 &    557 &                                 \{U2, Led Zeppelin\} \\
    41         &      40 &       8 &  0.211 &    999 &                          \{U2, Led Zeppelin, Queen\} \\
    42         &       6 &       5 &  0.242 &    991 &                       \{Elvis Presley, Johnny Cash\} \\
    43         &      35 &       7 &  0.246 &   1479 &  \{The Who, The Rolling Stones, The Kinks, The B... \\
    44         &      43 &      39 &  0.234 &   3404 &  \{The Who, The Rolling Stones, The Kinks, The B... \\
    45         &      44 &      10 &  0.236 &   3816 &  \{The Who, The Rolling Stones, The Kinks, The B... \\
    46         &      45 &      41 &  0.236 &   4815 &  \{The Who, The Rolling Stones, The Kinks, The B... \\
    47         &      17 &      15 &   0.283 &    663 &                      \{Lata Mangeshkar, Lead Belly\} \\
    48         &      37 &      13 &  0.318 &   2005 &  \{Ella Fitzgerald, Frank Sinatra, Dean Martin, ... \\
    49         &      46 &      42 &  0.337 &   5806 &  \{The Who, The Rolling Stones, The Kinks, The B... \\
    50         &      49 &      24 &  0.334 &   6046 &  \{The Who, The Rolling Stones, The Kinks, The B... \\
    51         &      48 &       9 &  0.379 &   2426 &  \{Ella Fitzgerald, Frank Sinatra, Dean Martin, ... \\
    52         &      51 &      47 &  0.408 &   3089 &  \{Ella Fitzgerald, Frank Sinatra, Dean Martin, ... \\
    53         &      52 &      50 &  0.414 &   9135 &  \{Ella Fitzgerald, Frank Sinatra, Dean Martin, ... \\
    54         &      53 &      18 &  0.468 &   9411 &  \{Ella Fitzgerald, Frank Sinatra, Dean Martin, ... \\
    55         &      54 &      26 &  0.481 &   9649 &  \{Ella Fitzgerald, Frank Sinatra, Dean Martin, ... \\
    56         &       1 &       0 &  0.487 &   1591 &                \{Ignacio Corsini, Francisco Canaro\} \\
    57         &      55 &      25 &  0.528 &   9889 &  \{Ella Fitzgerald, Frank Sinatra, Dean Martin, ... \\
    58         &      57 &      56 &  0.598 &  11480 &  \{Ella Fitzgerald, Frank Sinatra, Dean Martin, ... \\
    \bottomrule
    \end{tabular}
    \caption{\textbf{Linkage Matrix}. The table shows the formations of clusters, chronologically ordered. Each cluster is associated with two children clusters and a distance between children clusters. The \emph{size} represents the number of instances in the cluster. The \emph{qualitative values} shows the list of the unique qualitative values representative of the cluster.}
    \label{tab:linkage}
\end{table*}

In this section, we showcase the method HQC, and how it can be used to extract insights from the data. We applied HQC to two different problems. The first problem is the clustering of music artists, given a dataset with songs. The second use case is the clustering of company industries, given a dataset with yearly features for stocks. The code and data used for these experiments are available on the public repository \url{https://github.com/diogoseca/qualitative-clustering}.

The experiments in this section were performed using the following hardware: Intel® Core™ i7-4700HQ CPU @ 2.40GHz with 6 MB of cache, and 16 GB of DDR3L 1600 MHz SDRAM. The code was run on an Anaconda Python 3.7 environment under Ubuntu 20.20 LTS. In both uses cases presented, the clustering method proposed completed its computations in under 2 minutes.

\subsection{Clustering Musicians}

In this subsection, we showcase our method by identifying clusters of music artists using the context of thousands of quantified songs.

\begin{figure*}
    \centering
    \includegraphics[width=\textwidth]{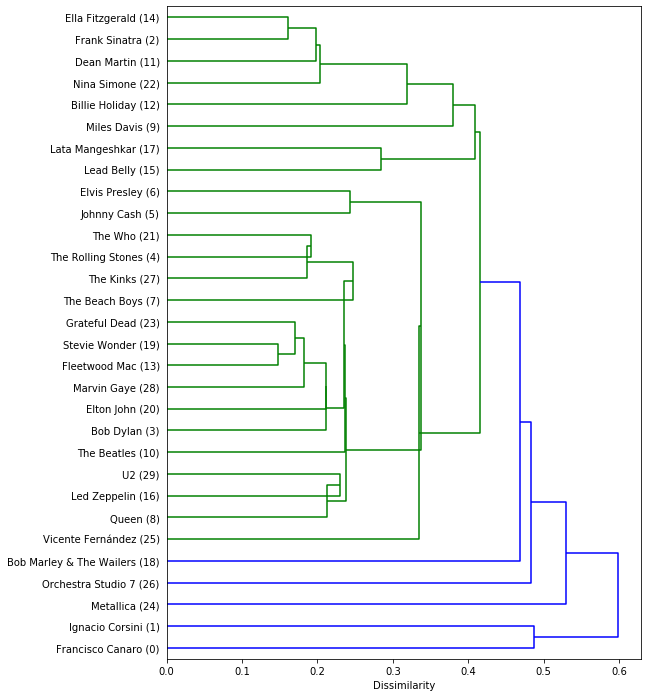}
    \caption{\textbf{Dendrogram for the qualitative variable \emph{artist}.} The left axis shows the quantitative value being clustered, i.e. the music artists, as well as its respective initial cluster id, between brackets. The bottom axis represents the dissimilarity, at which clusters where joined. Each connection represents a new cluster that was formed from joining previous two clusters.}
    \label{fig:dendrogram_artists}
\end{figure*}

\begin{figure*}
    \centering
    \includegraphics[width=\textwidth]{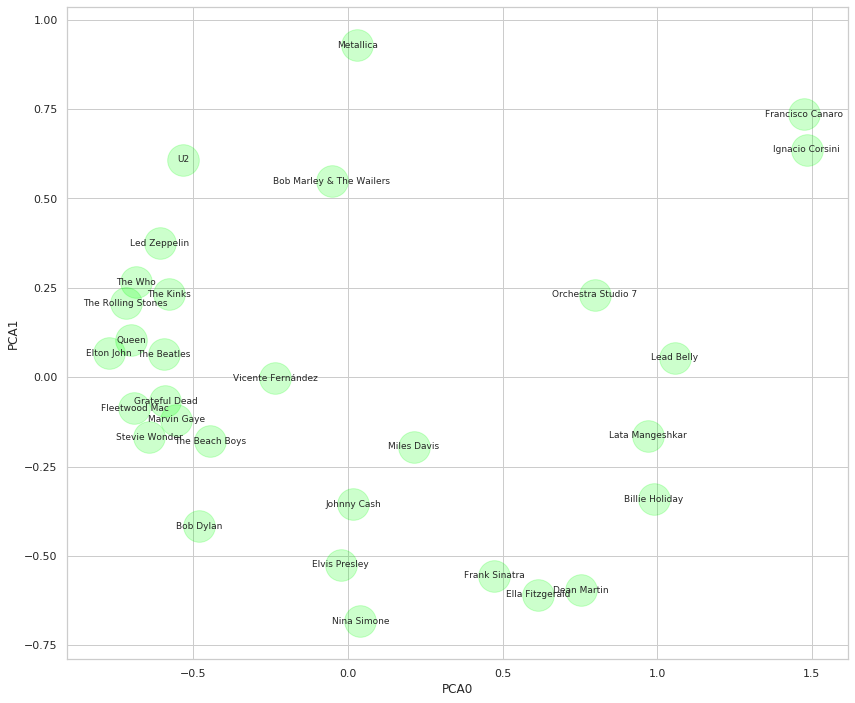}
    \caption{\textbf{Scatter plot of the qualitative variable in two dimensions.} The two components shown capture the most variance from the Dissimilarity Matrix.}
    \label{fig:pca2d}
\end{figure*}

The dataset used in this experiment is entitled \emph{Spotify Dataset 1921-2020} and is available on Kaggle \cite{spotify}. In order to simplify our demonstration, we only make use of instances from the main file \emph{data.csv} that contain one and only one author. The variable chosen for clustering was \emph{artists}. The quantitative variables chosen for providing the context were: \emph{acousticness}, \emph{danceability}, \emph{duration}\_ms, \emph{energy}, \emph{speechiness}, \emph{instrumentalness}, \emph{liveness}, \emph{loudness}, \emph{valence}, \emph{tempo}, \emph{popularity}, \emph{explicit}, \emph{mode}, and \emph{year}. In other words, the distributions from which we extracted the context were 14-dimensional. Moreover, we only considered the top 30 artists with the most songs in the dataset, namely: \emph{Francisco Canaro}, \emph{Ignacio Corsini}, \emph{Frank Sinatra}, \emph{Bob Dylan}, \emph{The Rolling Stones}, \emph{Johnny Cash}, \emph{Elvis Presley}, \emph{The Beach Boys}, \emph{Queen}, \emph{Miles Davis}, \emph{The Beatles}, \emph{Dean Martin}, \emph{Fleetwood Mac}, \emph{Billie Holiday}, \emph{Ella Fitzgerald}, \emph{Lead Belly}, \emph{Led Zeppelin}, \emph{Lata Mangeshkar}, \emph{Bob Marley \& The Wailers}, \emph{Stevie Wonder}, \emph{Elton John}, \emph{The Who}, \emph{Nina Simone}, \emph{Grateful Dead}, \emph{Vicente Fernández}, \emph{Metallica}, \emph{Orchestra Studio 7}, \emph{The Kinks}, \emph{Marvin Gaye}, and \emph{U2}. After this preselection, we are left with 11480 instances in the dataset, i.e. 11480 songs from 30 distinct music artists.

Commonly used visualizations such as distributions plots are feasible for a few qualitative values, in which each qualitative value is associated with a distinct color. In addition, these visualizations are done for a few quantitative dimensions. Otherwise, the plots become too large and complex. In cases of high dimensionality, such as the present use case, the visual tasks of Exploratory Data Analysis become complicated and confusing. To our advantage, our method is apt in dealing with high dimensionality.

Before being analyzed, the data is standardized. We start with 30 initial clusters, one for each artist. Initially, the instances in a cluster contain one and only one artist. Iteratively, clusters are paired up and form larger clusters. As the clusters increase in size to form clusters higher in the hierarchy, so too does the number of artists enclosed by the clusters.

After the dataset is processed by HQC, the resultant linkage matrix allows for the visualization of qualitative values in a dendrogram. The dendrogram allows the visualization of the connections and distances between clusters of qualitative values. The visualization of the dendrogram is done using the implementation by the Python package \emph{scipy}.

Interestingly, the dendrogram shows that progressive clusterings do not always result in larger distances between the remaining clusters. In some cases, the pairing of the minimum-distancing clusters $A$ with $B$ results in a cluster whose distance to another cluster $C$ could be lesser than the distance from $A$ to $B$.

The linkage matrix resultant from the clustering process can be seen in Table~\ref{tab:linkage}. Note that in the linkage matrix, the distances are not always increasing. Although this is natural using HAC, this is not guaranteed using HQC. In HQC, the process of joining two clusters could cause the resulting new cluster to become surprisingly similar to an already existing cluster. Notice the formation of cluster 35. The distance between \emph{The Kinks} and \emph{The Who} was 0.216, and the distance between \emph{The Kinks} and \emph{The Rolling Stones} was 0.201. However, the distance between \emph{The Kinks}, the cluster of both \emph{The Who} and \emph{The Rolling Stones} was only 0.186. It could be argued that, if the quantitative distributions accurately represent the tastes of a recreational music listener, and if such a listener liked \emph{The Who} and \emph{The Rolling Stones}, then he could appreciate the suggestion of listening to \emph{The Kinks}. This phenomenon can be seen in Figure~\ref{fig:dendrogram_artists}: between nodes 4, 21, and 27.

One alternative way for visualizing the dissimilarity of qualitative values is to apply Principal Component Analysis to the dissimilarity matrix and extract the two components which capture the most variance. Then, plot the qualitative values' components on a 2-dimensional scatter plot. This can be seen in Figure~\ref{fig:pca2d}. We can look at Figure~\ref{fig:pca2d} to better understand what is perhaps going on, in a more interpretable fashion. Indeed, \emph{The Kinks} seems to appear in the middle of both \emph{The Who} and \emph{The Rolling Stones}, w.r.t the first PCA component. 
Music artists are frequently classified in music genres. The classifications of music genre follow subjective and controversial criteria. Possibly due to the artists being associated with more than one genre, we can not find clusters of music artists that represent single music genres. 

The dissimilarity calculation can be used for grouping artists according to a given user's preference. In particular, the calculation of the dissimilarity between a pool of songs from a user's playlist and an artist the user has never listened to before should result in finding artists with songs that match the user's preference.


\subsection{Clustering Company Industries}

In this subsection, we showcase our method by identifying clusters of industries using the context of thousands of companies' financial statements.

\begin{figure*}
    \centering
    \includegraphics[width=\textwidth]{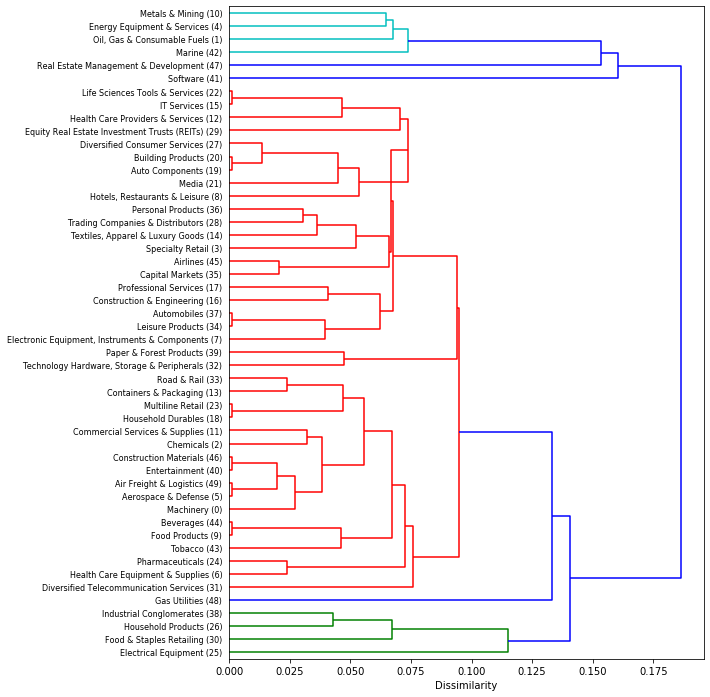}
    \caption{\textbf{Dendrogram for the qualitative variable \emph{industry}.} The left axis shows the quantitative value being clustered, i.e. the industry, as well as its respective initial cluster id, between brackets. The bottom axis represents the dissimilarity, at which clusters where joined. Each connection represents a new cluster that was formed from joining previous two clusters.}
    \label{fig:dendrogram_industries}
\end{figure*}

Publicly traded US companies have been classified using a taxonomy named Global Industry Classification Standard (GICS). In comparison to our statistical-based approach, the GICS hierarchical categorization uses a qualitative and subjective analysis. According to the Morgan Stanley Capital International (MSCI) and the Standard \& Poor's (S\&P), the GICS is "based on market-demand"~\cite{gics}.

Based on financial statements gathered from the Thomson Reuters Eikon API~\cite{eikon_2019} and on market data gathered from Yahoo Finance~\cite{yahoo_2019}, we have aggregated yearly features of several stocks into a working dataset. Moreover, each stock is annotated with the GICS sector, industry group, industry, and sub-industry. The dataset contains 16039 instances and 18 quantitative features, including the respective year.

In this use case, we will be using clustering the top 50 most frequent industries within in our dataset. More precisely, we will be using the unique values of the \emph{industry} variable for forming the initial clusters. After selecting filtering for the top 50 most frequent industries, the dataset is left with 15710 instances.

During the computation of the dissimilarity matrix, we notice that some comparisons between industries result in distances equal to zero, as seen in Figure~\ref{fig:dendrogram_industries}. In particular, this occurs: between \emph{Air Freight \& Logistics} and \emph{Aerospace \& Defense}, which belong to the sector of \emph{Industrials}; between \emph{Multiline Retail} and \emph{Household Durables}, which belong to the sector of \emph{Consumer Discretionary}; between \emph{Automobiles} and \emph{Leisure Products}, which belong to the sector of \emph{Consumer Discretionary}; and between \emph{Beverages} and \emph{Food Products}, which belong to the industry group of \emph{Food, Beverage \& Tobacco}, covered by the \emph{Consumer Staples sector}.
While the former zero distances may seem reasonable, given that the industries belong to the same sector, the following do not: \emph{Life Sciences Tools \& Services} belongs to the sector \emph{Health Care} while \emph{IT Services} belong to the sector\emph{Information Technology};
\emph{Auto Components} belongs to the sector \emph{Consumer Discretionary} while \emph{Building Products} belong to the sector \emph{Industrials}; \emph{Construction Materials} belongs to the sector \emph{Materials} while \emph{Entertainment} belong to the sector \emph{Communication Services}.
The previous seven comparisons were each performed using between 199 and 697 instances. The results sugests that these paired industries are highly correlated. Unless the results are spurious, the last 3 cases present three high correlations between industries, despite being from different sectors. 
Therefore, in the absence of additional empirical results, we caution the reader to be aware of these correlations when diversifying his investment portfolio. If the diversification is considered at the sector level, some of the assets chosen may actually end being highly correlated, despite being from distinct sectors. Instead, we caution the reader to consider diversifying his portfolio at the industry level.

\section{Conclusion}

Following a brief literature review and contextualization, we presented a novel method called Hierarchical qualitative Context Clustering (HQC). The method proposed can quantify dissimilarities between qualitative values based on the context of quantitative data. The Maximum Mean Discrepancy can compute these dissimilarities in a timely matter. The proposed method is especially valuable in cases where the number of unique qualitative values is high, or when there are many quantitative variables in the dataset. The visualization of HQC through dendrograms preserves the text information contained in the qualitative variables initially present in the datasets.

The experiment compared 30 music artists, using 14 continuous variables per song, with over 200 songs per artist. The method can also be used for creating recommendation systems for users who are interested in discovering new musical artists based on their preferences. Aside from clustering musical artists, clustering can also be applied to areas where the domain knowledge is unknown or partially known, e.g., financial assets, in which there is a lack of consensus for grouping investments.
\section{Open Research Questions}

This work further resulted in the following research questions. Firstly is to understand the context of a qualitative value not only by looking at both quantitative variables and qualitative variables. Secondly is to use the dissimilarity matrix as a starting point for encoding qualitative variables to quantitative variables. Finally is to test whether supervised learning improves using HQC clusters as features.
\end{multicols}

\bibliographystyle{unsrt}
\bibliography{references} 

\begin{thebibliography}{10}

\bibitem{shearer2000}
Colin Shearer.
\newblock The {CRISP-DM} model: the new blueprint for data mining.
\newblock {\em Int. J. Data Warehouse. Min.}, 5(4):13--22, 2000.

\bibitem{nogueira2014}
D~Nogueira.
\newblock Agile data mining: uma metodologia {\'a}gil para o desenvolvimento de
  projetos de data mining.
\newblock Master's thesis, Faculdade de Engenharia da Universidade do Porto.,
  2014.

\bibitem{hastie2009}
Trevor Hastie, Robert Tibshirani, and Jerome Friedman.
\newblock {\em The Elements of Statistical Learning: Data Mining, Inference,
  and Prediction}.
\newblock Springer, New York, NY, 2009.

\bibitem{velden2019}
Michel van~de Velden, Alfonso Iodice~D'Enza, and Angelos Markos.
\newblock Distance‐based clustering of mixed data.
\newblock {\em WIREs Comput Stat}, 11(3):e1456, May 2019.

\bibitem{rezankova2009}
Hana {\v R}ezankov{\'a} and B~Everitt.
\newblock Cluster analysis and categorical data.
\newblock {\em Statistika}, 89(2):216--232, 2009.

\bibitem{fayyad1993}
Usama~M Fayyad and Keki~B Irani.
\newblock {Multi-Interval} discretization of {Continuous-Valued} attributes for
  classification learning.
\newblock {\em IJCAI}, 1993.

\bibitem{huang1997}
Z~Huang.
\newblock Clustering large data sets with mixed numeric and categorical values.
\newblock In {\em Proceedings Of 1st {Pacific-Asia} Conference on Knowledge
  Discovery And Data Mining}, 1997.

\bibitem{domingos2012}
Pedro Domingos.
\newblock A few useful things to know about machine learning.
\newblock {\em Commun. ACM}, 55(10):78--87, October 2012.

\bibitem{gower1971}
J~C Gower.
\newblock A general coefficient of similarity and some of its properties.
\newblock {\em Biometrics}, 27(4):857--871, 1971.

\bibitem{huang1998}
Zhexue Huang.
\newblock Extensions to the k-means algorithm for clustering large data sets
  with categorical values.
\newblock {\em Data Min. Knowl. Discov.}, 2(3):283--304, September 1998.

\bibitem{andritsos2004}
Periklis Andritsos, Panayiotis Tsaparas, Ren{\'e}e~J Miller, and Kenneth~C
  Sevcik.
\newblock Limbo: Scalable clustering of categorical data.
\newblock In {\em Advances in Database Technology - {EDBT} 2004}, volume~29,
  pages 123--146. Springer Berlin Heidelberg, 2004.

\bibitem{kimes2017}
Patrick~K Kimes, Yufeng Liu, David Neil~Hayes, and James~Stephen Marron.
\newblock Statistical significance for hierarchical clustering.
\newblock {\em Biometrics}, 73(3):811--821, September 2017.

\bibitem{alamuri2014}
Madhavi Alamuri, Bapi~Raju Surampudi, and Atul Negi.
\newblock A survey of distance/similarity measures for categorical data, 2014.

\bibitem{le2005}
Si~Quang Le and Tu~Bao Ho.
\newblock An association-based dissimilarity measure for categorical data.
\newblock {\em Pattern Recognit. Lett.}, 26(16):2549--2557, December 2005.

\bibitem{szekely2014}
G{\'a}bor~J Sz{\'e}kely and Maria~L Rizzo.
\newblock Partial distance correlation with methods for dissimilarities.
\newblock {\em Ann. Stat.}, 42(6):2382--2412, December 2014.

\bibitem{hardoon2004}
David~R Hardoon, Sandor Szedmak, and John Shawe-Taylor.
\newblock Canonical correlation analysis: an overview with application to
  learning methods.
\newblock {\em Neural Comput.}, 16(12):2639--2664, December 2004.

\bibitem{robert1976}
Paul Robert and Yves Escoufier.
\newblock A unifying tool for linear multivariate statistical methods: the
  {RV-coefficient}.
\newblock {\em J. R. Stat. Soc. Ser. C Appl. Stat.}, 25(3):257--265, 1976.

\bibitem{ienco2012}
Dino Ienco, Ruggero~G Pensa, and Rosa Meo.
\newblock From context to distance: Learning dissimilarity for categorical data
  clustering.
\newblock {\em ACM Trans. Knowl. Discov. Data}, 6(1):1:1--1:25, March 2012.

\bibitem{khorshidpour2010}
Z~Khorshidpour, S~Hashemi, and A~Hamzeh.
\newblock Distance learning for categorical attribute based on context
  information.
\newblock In {\em 2010 2nd International Conference on Software Technology and
  Engineering}, volume~2, pages V2--296--V2--300, October 2010.

\bibitem{jorge2006}
Al{\'\i}pio~M Jorge, Paulo~J Azevedo, and Fernando Pereira.
\newblock Distribution rules with numeric attributes of interest, 2006.

\bibitem{lucas2007}
Joel~P Lucas, Al{\'\i}pio~M Jorge, Fernando Pereira, Ana~M Pernas, and Amauri~A
  Machado.
\newblock A tool for interactive subgroup discovery using distribution rules.
\newblock {\em Progress in Artificial Intelligence}, pages 426--436, 2007.

\bibitem{engmann2011}
S~Engmann and D~Cousineau.
\newblock Comparing distributions: the two-sample {Anderson-Darling} test as an
  alternative to the {Kolmogorov-Smirnoff} test.
\newblock {\em Journal of applied quantitative}, 2011.

\bibitem{pettitt1976}
A~N Pettitt.
\newblock A {Two-Sample} {Anderson--Darling} rank statistic.
\newblock {\em Biometrika}, 63(1):161--168, 1976.

\bibitem{baksajev2015}
Aleksej Bak{\v s}ajev and Rimantas Rudzkis.
\newblock Multivariate goodness of fit tests based on the kernel density
  estimators.
\newblock {\em Nonlinear Anal. Model. Control}, pages 585--602, 2015.

\bibitem{rosenblatt1952}
Murray Rosenblatt.
\newblock Remarks on a multivariate transformation.
\newblock {\em Ann. Math. Stat.}, 23(3):470--472, September 1952.

\bibitem{gretton2012}
Arthur Gretton, Karsten~M Borgwardt, Malte~J Rasch, Bernhard Sch{\"o}lkopf, and
  Alexander Smola.
\newblock A kernel {Two-Sample} test.
\newblock {\em J. Mach. Learn. Res.}, 13(25):723--773, 2012.

\bibitem{franca2017}
Guilherme Fran{\c c}a, Maria~L Rizzo, and Joshua~T Vogelstein.
\newblock Kernel k-groups via hartigan's method.
\newblock {\em arXiv e-prints}, page arXiv:1710.09859, October 2017.

\bibitem{arcones1992}
Miguel~A Arcones and Evarist Gine.
\newblock On the bootstrap of {U} and {V} statistics.
\newblock {\em Ann. Stat.}, pages 655--674, 1992.

\bibitem{kim2016}
Been Kim, Rajiv Khanna, and Oluwasanmi~O Koyejo.
\newblock Examples are not enough, learn to criticize! criticism for
  interpretability.
\newblock In D~D Lee, M~Sugiyama, U~V Luxburg, I~Guyon, and R~Garnett, editors,
  {\em Advances in Neural Information Processing Systems 29}, pages 2280--2288.
  Curran Associates, Inc., 2016.

\bibitem{spotify}
Yamaç~Eren Ay.
\newblock Spotify dataset 1921-2020, 160k+ tracks.
\newblock
  \url{https://www.kaggle.com/yamaerenay/spotify-dataset-19212020-160k-tracks},
  5 2020.

\bibitem{gics}
{MSCI}.
\newblock {GICS} - global industry classification standard - {MSCI}.
\newblock \url{https://www.msci.com/gics}.
\newblock Accessed: 2020-6-12.

\bibitem{eikon_2019}
{Thomson Reuters}.
\newblock {Eikon API}.
\newblock Subscription Service.
\newblock Accessed in February, 2019.

\bibitem{yahoo_2019}
{Yahoo}.
\newblock {Yahoo Finance}.
\newblock \url{https://finance.yahoo.com/}.
\newblock Accessed in February, 2019.

\end{thebibliography}
\end{document}